 \documentclass[pmlr,twocolumn,wcp]{jmlr} 

\usepackage{booktabs}
\usepackage[load-configurations=version-1]{siunitx} 

\jmlrvolume{ML4H Extended Abstract arXiv Index}
\jmlryear{}
\jmlrsubmitted{2020}
\jmlrpublished{}
\jmlrworkshop{Machine Learning for Health (ML4H) Workshop at NeurIPS 2020} 

\title[Detecting small polyps using a Dynamic SSD-GAN]{Detecting small polyps using a Dynamic SSD-GAN}

\author{%
\Name{Daniel C. Ohrenstein}\nametag{\textsuperscript{\dag}} \Email{daniel.ohrenstein.19@ucl.ac.uk}\\
\Name{Patrick Brandao}\nametag{\textsuperscript{\S}} \Email{patrickbrandao@odin-vision.com}\\
\Name{Daniel Toth}\nametag{\textsuperscript{\S}} \Email{danieltoth@odin-vision.com}\\
\Name{Laurence Lovat}\nametag{\textsuperscript{\ddag}} \Email{l.lovat@ucl.ac.uk}\\
\Name{Danail Stoyanov}\nametag{\textsuperscript{\ddag}} \Email{danail.stoyanov@ucl.ac.uk} \\
\Name{Peter Mountney}\nametag{\textsuperscript{\S}} \Email{petermountney@odin-vision.com}\\
\addr \textsuperscript{\dag}Department of Computer Science \\
University College London \\
London, UK \\
\textsuperscript{\S}Odin Vision \\
London, UK \\
\textsuperscript{\ddag}Wellcome/EPSRC Centre for Interventional and Surgical Sciences (WEISS) \\
University College London \\
London, UK
 }

\begin{document}

\maketitle

\begin{abstract}
  Endoscopic examinations are used to inspect the throat, stomach and bowel for polyps which could develop into cancer. Machine learning systems can be trained to process colonoscopy images and detect polyps. However, these systems tend to perform poorly on objects which appear visually small in the images. It is shown here that combining the single-shot detector as a region proposal network with an adversarially-trained generator to upsample small region proposals can significantly improve the detection of visually-small polyps. The Dynamic SSD-GAN pipeline introduced in this paper achieved a 12\% increase in sensitivity on visually-small polyps compared to a conventional FCN baseline.
\end{abstract}
\begin{keywords}
Polyp Detection, CNNs, GANs
\end{keywords}

\section{Introduction}
\label{sec:intro}

\begin{figure*}[bt]
    \centering
    \includegraphics[width=\linewidth]{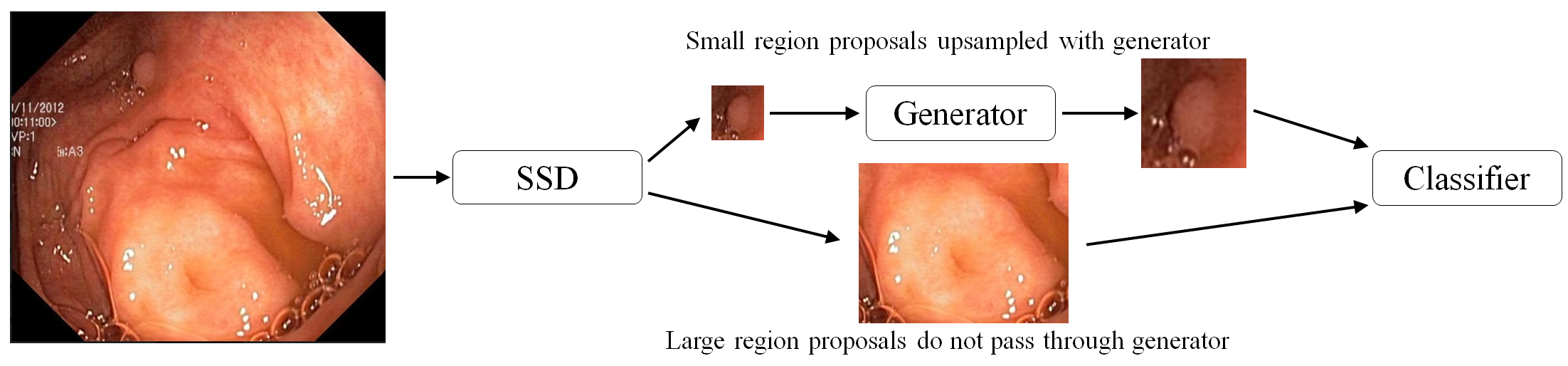}
    \caption{Pipeline for Dynamic SSD-GAN detector. Large region proposals do not pass through the generator and are instead simply reshaped (using bicubic interpolation where necessary) to the classifier input size.}
    \label{fig:dynamic_pipeline}
\end{figure*}

Detecting polyps during endoscopic procedures is challenging and 25\% of polyps are missed \citep{kumar2017adenoma}. It is believed that a 1\% increase in adenoma (polyp associated with a greater risk of cancer) detection rate (ADR) is associated with a 3\% decrease in the risk of interval cancer. \citep{corley2014adenoma}.  A number of technologies have arisen to assist endoscopists with their examinations. These technologies have largely focused on the endoscope itself \citep{g-eye, endocuff, endoring} but recently, AI-powered tools have been developed to aid polyp detection. Various groups \citep{park,nima,cad} have employed deep convolutional neural networks (CNNs) enabling real-time and detailed image analysis. More complicated models \citep{park, mahmood, qadir} can include time-series statistics such as hidden Markov models, taking advantage of the inherent correlation between consecutive frames. Like many object detection models, these systems have a tendency to perform poorly on visually-small objects \citep{pham2017evaluation}. Object size should be understood to mean size relative to the field of view. This paper introduces a novel approach to the task, adapting a technique previously used to identify small faces in images \citep{tiny_faces}. This approach is compared to a baseline model consisting of a fully-convolutional network model with a ResNet-101 backbone (FCN-ResNet101), selected as its ability to accurately detect polyps is well-established \citep{brandao2017fully}. The Dynamic SSD-GAN pipeline achieved a 10\% increase in sensitivity on visually-small polyps compared to the baseline.

\section{Dataset}
\label{section:dataset}

For training and testing, Version 2 of the Kvasir Dataset \citep{kvasir} was used. Since the aim of this investigation is to improve performance on small polyps specifically, it was important to ensure a large pool of data in this category. To simulate visually-small polyps, the polyp images were duplicated, scaled by 75\% and padded to their original sizes. The size of the ground truth bounding boxes (along the diagonal) were divided by the size of the full image to generate a set of relative object sizes for each image. These relative sizes were used to stratify the frames into groups for analysis.

\section{Method}

The Dynamic SSD-GAN pipeline consists of three stages:
\begin{itemize}
    \setlength{\itemsep}{0pt}
    \setlength{\parskip}{0pt}
    \item Use a single-shot detector (SSD) \citep{SSD} as a region proposal network.
    \item Dynamic step: Pass any small region proposals (both dimensions less than 200 pixels) through a generator, trained within a generative adversarial network framework \citep{goodfellow2014generative}, for upsampling and refinement.
    \item Classify all region proposals using a convolutional neural network (CNN).
\end{itemize}
The dynamic step recognises that not all the region proposals require upsampling. Reducing a region proposal to some small generator input size and subsequently upsampling constitutes a significant loss of data and is neither efficient nor useful. Therefore large region proposals are simply classified without generative upsampling, see Figure \ref{fig:dynamic_pipeline}.

\subsection{Implementation}

The SSD300 version of the single-shot detector \citep{SSD} was used with default prior scales and aspect ratios. A number of randomly-applied data augmentations (including saturation and contrast adjustments, zoom operations and flips) were used to facilitate effective training. Since the region proposals are subsequently processed and classified by further networks, the SSD was tuned be highly sensitive and over-detect polyps.

The loss function used to train the GAN effectively combines ideas from SR-GAN and Cycle-GAN \citep{super_resolution_gan,cyclegan}. The main adaptations made to the original architecture were reducing the upsampling factor from x4 to x2 (so the input size can be increased to better match the majority of small polyps and increase accuracy) and changing the input size from 32x32 to 150x150 so that the output size is large enough for accurate classification. The use case in the original paper is identifying extremely small faces in photos. Since the polyps in the Kvasir dataset are much larger than these faces, such a small input size would result in a significant loss of information from each region proposal. At test-time, the discriminator is discarded.

The classifier network uses the same architecture as the discriminator. It was trained on a dataset consisting of large polyps as well as generator-upsampled small polyps and negative regions from the same training set as was used to train the SSD (to avoid data leakage). Crucially, this dataset included authentic, computer-generated and negative data.

\subsection{Ablation Study}
The contribution of the generative modelling step was verified by constructing an equivalent pipeline with the generator replaced with bicubic interpolation. These were tested on the 75\%-scaled images in order to give a fair balance between regions that are above and below the dynamic step threshold for use of the generator.

\begin{figure}[htbp]
\floatconts
{fig:dynamic_stratified_sensitivity}
  {\caption{Performance of best FCN-ResNet101 and Dynamic SSD-GAN. The images have been stratified according to relative size of polyp. Test set: 400 positive, 400 negative, unscaled images.}}
  {%
    \subfigure[Baseline.]{\label{fig:baseline}%
     \includegraphics[width=0.95\linewidth]{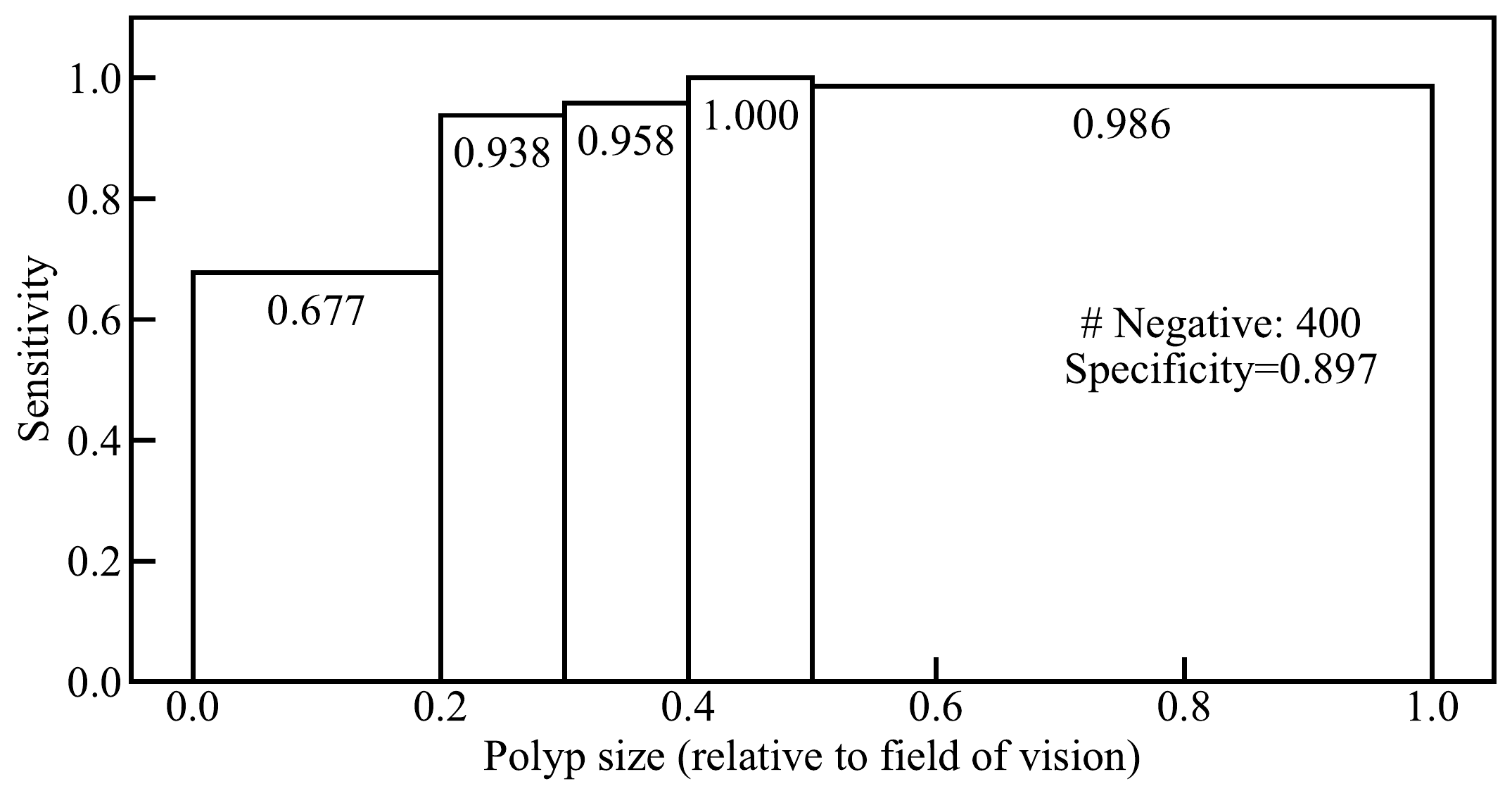}}%
     \qquad
    \subfigure[Dynamic SSD-GAN.]{\label{fig:dynamic_ssd_gan}%
      \includegraphics[width=0.95\linewidth]{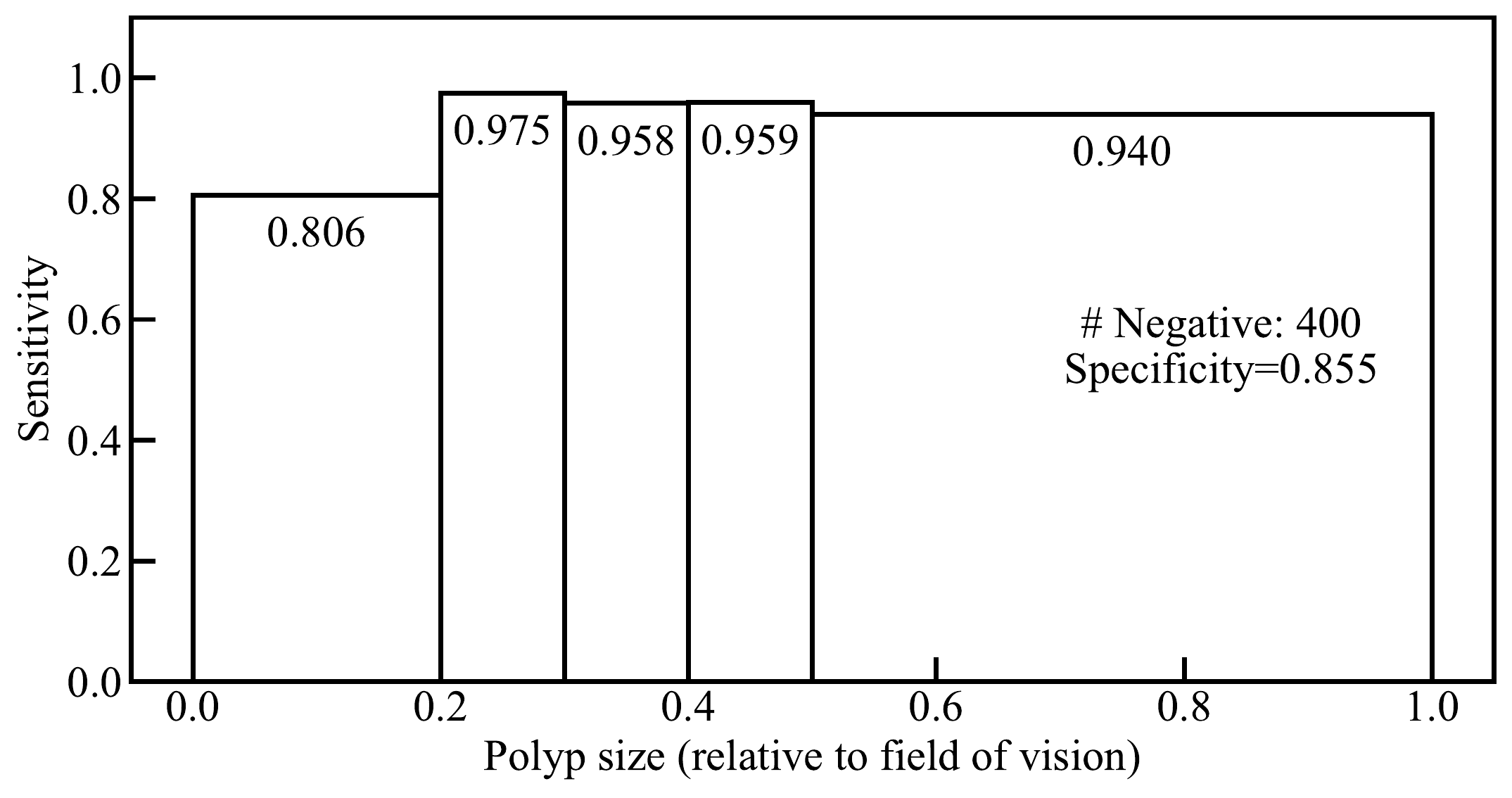}}
  }
  
\floatconts
{fig:occlusion}
  {\caption{Dynamic SSD-GAN identified difficult and visually-small objects which the baseline failed to detect.}}
  {%
  \includegraphics[width=0.95\linewidth]{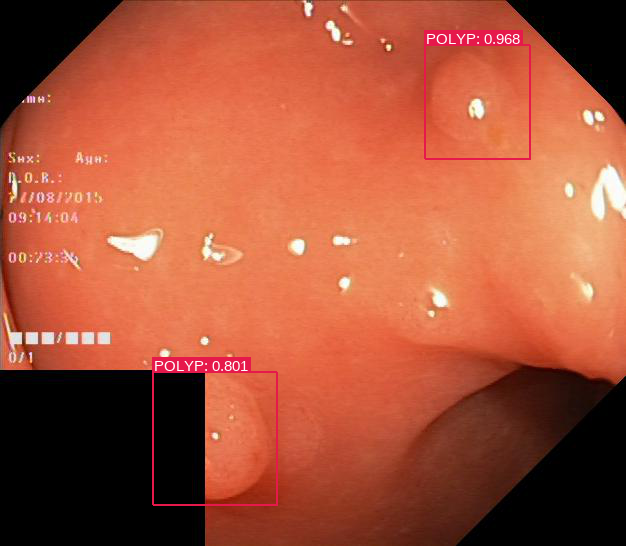}
  }
\end{figure}

\begin{table*}[hthb]
    \centering
    \caption{Summary of model performance (same models as in Figure \ref{fig:dynamic_stratified_sensitivity}). An overall sensitivity improvement of 11\% was observed. Test set: 400 positive, 400 negative, unscaled images. (TP=True Positives, FP=False Positives, TN=True Negatives, FN=False Negatives.)}
    \label{tab:model_eval}
    \begin{tabular}{c c c c c c c}
        \toprule
        Detector & Sensitivity & Specificity & TP & FP & TN & FN\\
        \midrule
        FCN-ResNet101 & 0.823 & 0.897 & 344 & 44 & 385 & 74 \\
        Dynamic SSD-GAN & 0.935 & 0.855 & 391 & 60 & 354 & 27 \\
        \bottomrule
    \end{tabular}

    \centering
    \caption{Ablation study: Comparison of Dynamic SSD-GAN performance with and without a learned deep generative model to upsample the small region proposals. Test set: 400 positive, 400 negative, 75\%-scaled images.}
    \label{tab:ablation}
    \begin{tabular}{c c c c c c c}
        \toprule
        Upsampling Method & Sensitivity & Specificity & TP & FP & TN & FN\\
        \midrule
        Generator & 0.909 & 0.808 & 380 & 79 & 333 & 38\\
        Bicubic Interpolation & 0.816 & 0.805 & 341 & 80 & 331 & 77 \\
        \bottomrule
    \end{tabular}
\end{table*}

\section{Results}

Figure  \ref{fig:dynamic_stratified_sensitivity} shows the performance of the detectors stratified according to relative polyp size. The size bands were selected to divide the images approximately evenly. As expected, there is a large drop-off in performance of the baseline model as the relative size decreases. Against the visually smallest category of polyps, the Dynamic SSD-GAN detector maintains a notably larger sensitivity than the baseline, without compromising performance on larger objects. Figure \ref{fig:occlusion} shows an image from the Kvasir dataset containing two difficult-to-detect polyps. One is significantly occluded (around 50\%) and the other is subtly textured and flat. The Dynamic SSD-GAN detector was able to detect both polyps with tight bounding boxes and high confidence scores, whereas the baseline failed to detect either. Table \ref{tab:model_eval} summarises the performance of the two detectors. The Dynamic SSD-GAN detector misses far fewer polyps than the baseline, although a small increase in the number of false positives was noted. The results of the ablation study are shown in Table \ref{tab:ablation}. The considerable increase in sensitivity when using the generator compared to bicubic interpolation shows that deep generative modelling provides a dominant contribution to the overall performance of the Dynamic SSD-GAN.

\section{Conclusion}
Notable improvements in small polyp detection rates can be achieved using generative adversarial networks. By incorporating a generator which can accurately super-resolve small region proposals, consistently high sensitivity was maintained across all relative polyp sizes in the Kvasir dataset. A key point of note is that the input size and upsampling factors of the generator need to be chosen extremely carefully. If the input size is too small, there will be significant data loss from many region proposals. There is potential to develop these ideas further, with detectors containing a filter to separate region proposals into size categories, followed by size-specific generative upsampling.


\acks{Concepts and information presented are based on research and are not commercially available. Due to regulatory reasons, the future availability cannot be guaranteed. This work is supported, in part, by InnovateUK 26673.}


\bibliography{references}

 \end{document}